# Nodule-DETR: A Novel DETR Architecture with Frequency-Channel Attention for Ultrasound Thyroid Nodule Detection

Jingjing Wang[1], Qianglin Liu[2], Zhuo Xiao[1], Xinning Yao[1], Bo Liu[1,3,*], Lu Li[1], Lijuan Niu[2], Fugen Zhou[1,3]

[1] *Image Processing Center, Beihang University, Beijing 100191, People's Republic of China*

[2] *Department of Ultrasound, National Cancer Center/National Clinical Research Center for Cancer/Cancer Hospital, Chinese Academy of Medical Sciences and Peking Union Medical College, Beijing, 100021, People's Republic of China*

[3] *Beijing Advanced Innovation Center for Biomedical Engineering, Beihang University, Beijing 100083, People's Republic of China*

## ABSTRACT

Thyroid cancer is the most common endocrine malignancy, and its incidence is rising globally. While ultrasound is the preferred imaging modality for detecting thyroid nodules, its diagnostic accuracy is often limited by challenges such as low image contrast and blurred nodule boundaries. To address these issues, we propose Nodule-DETR, a novel detection transformer (DETR) architecture designed for robust thyroid nodule detection in ultrasound images. Nodule-DETR introduces three key innovations: a Multi-Spectral Frequency-domain Channel Attention (MSFCA) module that leverages frequency analysis to enhance features of low-contrast nodules; a Hierarchical Feature Fusion (HFF) module for efficient multi-scale integration; and Multi-Scale Deformable Attention (MSDA) to flexibly capture small and irregularly shaped nodules. We conducted extensive experiments on a clinical dataset of real-world thyroid ultrasound images. The results demonstrate that Nodule-DETR achieves state-of-the-art performance, outperforming the baseline model by a significant margin of 0.149 in mAP@0.5:0.95. The superior accuracy of Nodule-DETR highlights its significant potential for clinical application as an effective tool in computer-aided thyroid diagnosis. The code of work is available at https://github.com/wjj1wjj/Nodule-DETR.

## 1. Introduction

Thyroid nodules are a common endocrine disease [1], broadly classified as benign or malignant [2]. The global incidence of thyroid cancer has been significantly rising [3]. While the majority of thyroid cancers have a positive prognosis, some malignant nodules are still at risk of progression. Early diagnosis and timely treatment are crucial for preventing thyroid cancer progression and reducing mortality rates, making effective thyroid nodule detection methods of great clinical importance [4].

Current diagnostic methods include ultrasound (US), computed tomography (CT), magnetic resonance imaging (MRI), fine-needle aspiration (FNA), and physical examination. Among these, ultrasound imaging is preferred due to its efficiency, non-invasiveness, and affordability [5]. Diagnosis via ultrasound relies on morphological features like shape, boundary definition, and internal echoes to determine benign or malignant status. However, radiologists typically diagnose nodules based on visual assessment of sonographic features, which is relatively subjective and relies heavily on the individual radiologist's clinical experience. Consequently, developing accurate methods for automated and objective detection and classification of thyroid nodules is critically important.

With advancements in artificial intelligence, computer-aided diagnosis (CAD) systems based on deep learning have become a mainstream approach in thyroid nodule diagnosis, aiming to reduce the workload of physicians and improve diagnostic accuracy [6]. These systems are primarily built on two main architectures with a fundamental trade-off. CNN-based methods, such as YOLO [7] [8] [9] and Faster R-CNN [10] [11] [12], use convolutional kernels that excel at extracting local features but are limited in their ability to perceive long-range dependencies [13] [14]. Transformer-based methods, in contrast, are designed for global modeling but are less adept at capturing fine-grained local details [15] [16]. To resolve this trade-off, recent work has focused on DETR (DEtection TRansformer) [17], a hybrid architecture that combines a CNN backbone with a Transformer. DETR reframes object detection as a direct set prediction problem, creating an end-to-end framework that eliminates the need for hand-designed components like anchors [18] and non-maximum suppression (NMS) [19]. The success of this architecture has led to its growing adoption in medical image analysis [20] [21] [22] and, specifically, for thyroid nodule diagnosis [23] [15].

Despite advancements in computer-aided detection (CAD) systems for ultrasound thyroid nodule analysis, significant challenges remain. A key challenge is that many detection algorithms rely on clearly defined nodule boundaries and distinct features to achieve accurate identification. However, thyroid nodules in ultrasound images often exhibit blurred boundaries, irregular shapes, and characteristics that closely resemble surrounding normal tissue, which significantly impairs detection accuracy [24]. Additionally, many existing methods process images solely in the spatial domain, making them less effective at handling common ultrasound artifacts, such as speckle noise and size limitations [25]. These spatial-domain approaches also struggle to efficiently compress the substantial redundant information in medical images during initial processing, further reducing their robustness.

To address these challenges, we propose Nodule-DETR, a novel model based on the DETR framework that uses Multi-Spectral Frequency Channel Attention (MSFCA) to improve diagnostic accuracy and efficiency. The model's core strategy is to leverage frequency analysis to decomposes images into different components, which allows the model to use high frequencies to highlight fine details like nodule boundaries and internal structure, while using low frequencies to capture the overall morphology, thereby enhancing feature representation. The Nodule-DETR architecture is further enhanced with two key enhancements. A Hierarchical Feature Fusion (HFF) module is designed to achieve more efficient multi-scale feature fusion, preserving both high-resolution details and rich semantic context. Additionally, Multi-Scale Deformable Attention (MSDA) is incorporated to improve the model's ability to flexibly detect nodules that are small or irregularly shaped.

The main contributions are summarized as follows:

1. A novel model, Nodule-DETR, is proposed, which demonstrates enhanced localization of thyroid nodules in ultrasound images and achieves superior detection performance compared to many existing methods.

2. To capture the diverse characteristics of thyroid nodules and make up for the neglected valuable information, a Multi-Spectral Frequency Channel Attention is introduced to enhance the model's capability in detecting low-contrast and blurred-boundary nodules.

3. A Hierarchical Feature Fusion module is designed into the model, serving as an effective multi-scale feature fusion block across feature hierarchies, further boosting detection accuracy.

4. To improve the model's robustness in detecting irregularly shaped nodules, a Multi-Scale Deformable Attention is incorporated which enhances detection robustness and accuracy.

The remainder of this paper is organized as follows: Section 2 reviews related work on thyroid nodule detection. Section 3 details the network's overall framework and the architecture of the proposed modules, including MSFCA, HFF, and MSDA. Section 4 presents the numerical experiments of the proposed model. Section 5 concludes the paper and discusses future directions.

## 2. Related works

Medical image object detection is a critical research area that significantly aids in clinical diagnosis. Previous work in this field, particularly for thyroid nodule diagnosis, has been dominated by three main architectural paradigms: CNN-based methods, Transformer-based methods, and hybrid approaches that combine the two.

### 2.1. CNN-based methods

CNNs have long been foundational for feature extraction in ultrasound images and are widely used in thyroid nodule detection. Object detection algorithms based on CNNs are primarily categorized into two main approaches: one-stage and two-stage algorithms. In one-stage object detection algorithms, the generation of candidate regions and category prediction are unified into a single step. Classic examples include YOLO [26], SSD [27], and RetinaNet [18]. For instance, Wu et al. [8] enhanced the YOLO model by leveraging thyroid video frame characteristics and introduced Cache-Track, a post-processing method that refines detection results by propagating them across adjacent frames using contextual relationships. Song et al. [28] improved the SSD network by incorporating a spatial pyramid network, developing a multi-scale SSD for thyroid nodule detection that combines global and local information. TUN-Det [29], a RetinaNet-like nodule detection model, utilizes a multi-head design to embed an ensemble strategy within an end-to-end module, improving accuracy and robustness by fusing multiple outputs generated by diverse sub-modules.

In contrast, two-stage object detection algorithms first generate candidate regions and then perform classification and precise localization. Representative algorithms for this approach include RCNN [30], SPP-Net [31], Fast RCNN [32] and Faster RCNN [33]. For example, Liu et al. [34] proposed a prior knowledge-guided nodule detection model based on Faster RCNN that enhances detection accuracy via multi-scale feature fusion. Abdolali et al. [35] developed an innovative deep neural network based on Mask R-CNN, which achieved strong performance in thyroid nodule detection by optimizing bounding box regression and mask generation with a regularized loss function.

Despite these advances, CNNs possess an inherent limitation. Their reliance on local convolution operations restricts their ability to capture long-range dependencies [13] [14]. This is a significant drawback when analyzing thyroid nodules, which exhibit substantial variability in size, shape, and boundary definition, as the fixed, local receptive fields of CNNs often fail to capture these complex morphological features [36] [37].

**2.2. Transformer-based methods**

To address the challenge of capturing global context, researchers turned to the Transformer[37] architecture, which was originally developed for natural language processing. Transformers utilize a self-attention mechanism to model long-range dependencies, giving them a strong global modeling capability that CNNs lack. Vision Transformer (ViT) [38] successfully applied this architecture to computer vision by treating image patches as a sequence, preserving spatial information with positional embeddings. This has led to the application of Transformers in various thyroid imaging tasks, such as nodule segmentation and classification [39] [40] [41].

While the methods discussed above have achieved great success, they are not well-suited for thyroid nodule detection for the following reasons. Firstly, by flattening image patches into vectors, ViT can lose critical local feature information, such as edge details and internal textures, which is crucial for accurately detecting small nodules [16]. Secondly, although ViT uses positional encodings, its ability to represent precise spatial locations and relationships can be less accurate than the inherent spatial preservation of CNNs, which is critical for differentiating nodules from surrounding tissue [15].

**2.3. The Hybrid Solution: DETR**

To overcome the respective limitations of pure CNN and Transformer models, DETR [17] was developed as a hybrid model that combines a CNN backbone with a Transformer encoder-decoder. DETR reframes object detection as a direct set prediction problem, which creates a simplified, end-to-end pipeline by eliminating the need for hand-designed components like anchor boxes [18] and non-maximum suppression (NMS) [19]. Ramezani et al. [42] employed the DETR framework for sparse lung nodule detection and addressed data imbalance by introducing a customized focal loss function. Phu et al. [43] applied the DETR model to SPECT images for evaluating remaining thyroid tissues following thyroidectomy. Zhang et al. [23] improved upon the DETR architecture by designing a parallel decoder structure tailored for multi-label thyroid nodule classification in ultrasound image, thereby enhancing classification accuracy. Zhou et al. [15] proposed Thyroid-DETR for thyroid nodule detection, which embeds self-attention into the backbone network to effectively capture the diverse features of early-stage thyroid nodules.

Following a review of DETR-series models, including DAB-DETR [44] and Deformable-DETR [45], we selected DN-DETR [46], as our baseline due to its superior performance, which is attributed to its novel query denoising training method.

# 3. Methods

**3.1. Overall framework**

The detection of thyroid nodules in ultrasound imaging is inherently challenging due to factors like low image contrast, blurred nodule boundaries, and the significant variation in nodule size and shape. To overcome these obstacles, we propose Nodule-DETR, a novel end-to-end object detection architecture. As illustrated in Figure 1, our framework is built upon the robust foundation of DN-DETR[46] but incorporates three specialized modules—MSFCA, HFF, and MSDA—each designed to address a specific challenge in thyroid nodule detection.

The architectural data flow begins with the backbone network, which is responsible for initial feature extraction. We employ a ResNet50 backbone, but with a critical enhancement: the integration of the MSFCA module within its residual blocks (Section 3.2). By operating in the frequency domain, MSFCA is uniquely equipped to enhance features of low-contrast and boundary-blurred nodules, which are often indistinct in the spatial domain alone. This allows the backbone to produce feature maps that are more discriminative from the outset.

Following the backbone, the extracted multi-scale feature maps are processed by a dedicated Hierarchical Feature Fusion module (Section 3.3). This module serves as a bridge between the backbone and the Transformer encoder. Its purpose is to intelligently fuse the feature maps from different network depths into a cohesive and powerful multi-scale feature pyramid. By using lightweight convolutions and a down-top fusion pathway, HFF efficiently preserves the high-resolution spatial details from shallow layers (vital for small nodules) while enriching them with the high-level semantic context from deeper layers.

The resulting multi-scale feature maps is then passed to the core of our detection model: the Transformer encoder-decoder. Here, we introduce our final innovation by replacing the standard attention mechanism with Multi-Scale Deformable Attention (Section 3.4). Unlike standard attention, which computes dense relationships across the entire feature map, MSDA adaptively samples a small set of key points around a reference point. This makes it exceptionally adept at focusing on small or irregularly shaped objects, significantly improving detection accuracy and convergence speed. The MSDA-enhanced encoder processes the global context of the feature maps, and the decoder uses these enriched features to perform set prediction.

Finally, the decoder's output is fed into prediction heads, which consist of Feed-Forward Networks that generate the final bounding box coordinates and class labels for each detected nodule. By synergistically combining these specialized components, the Nodule-DETR framework provides a comprehensive solution tailored to the unique challenges of ultrasound-based thyroid nodule detection.

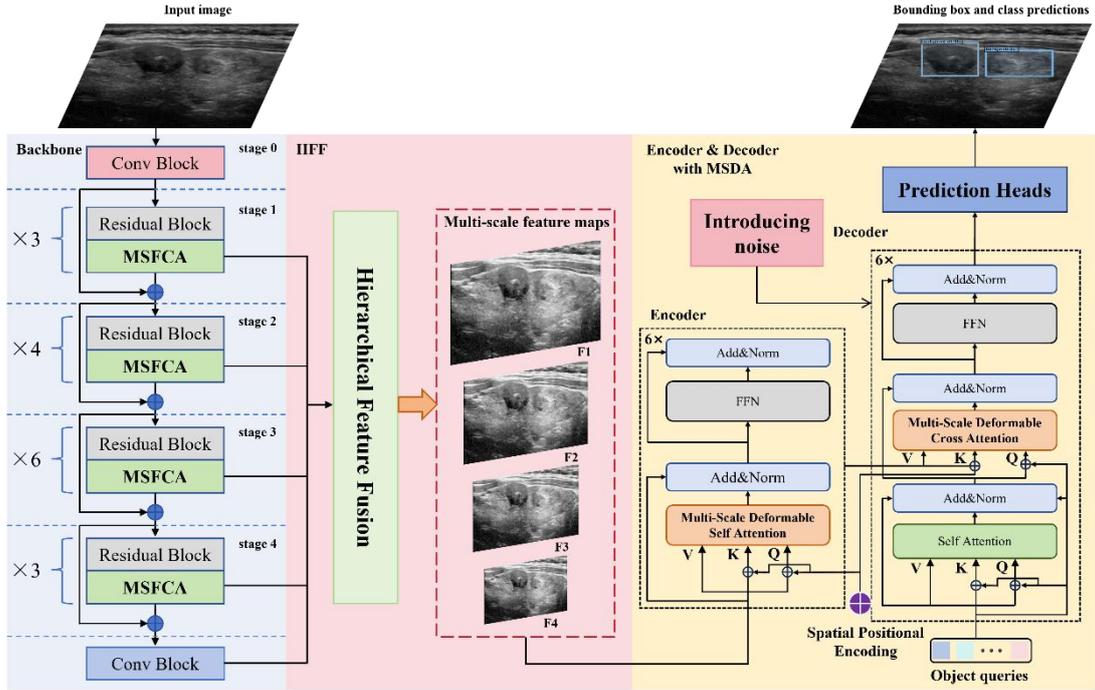

Figure. 1. Overall architecture of the Nodule-DETR detector.

## 3.2. Multi-Spectral Frequency Channel Attention (MSFCA) Module

Ultrasound images often suffer from low contrast and noise, limiting spatial-domain attention mechanisms (e.g., SENet[47], CBAM[48]) in distinguishing subtle nodule features. Frequency-domain analysis, however, can reveal critical information, as different frequency components correspond to distinct image characteristics (e.g., high frequencies for edges/textures, low frequencies for global structure). To leverage this, we introduce the Multi-Spectral Frequency Channel Attention module, designed to enhance detection of low-contrast, boundary-blurred nodules by focusing on discriminative frequency components. As illustrated in Figure 2, a comparative visualization shows that feature maps from the MSFCA-equipped model demonstrate more focused attention on the nodule region, whereas the baseline model's attention is scattered. The progressively darker red shades in the heatmaps, indicating heightened model attention, visually confirm MSFCA's focus on the nodule region within specific frequency bands.

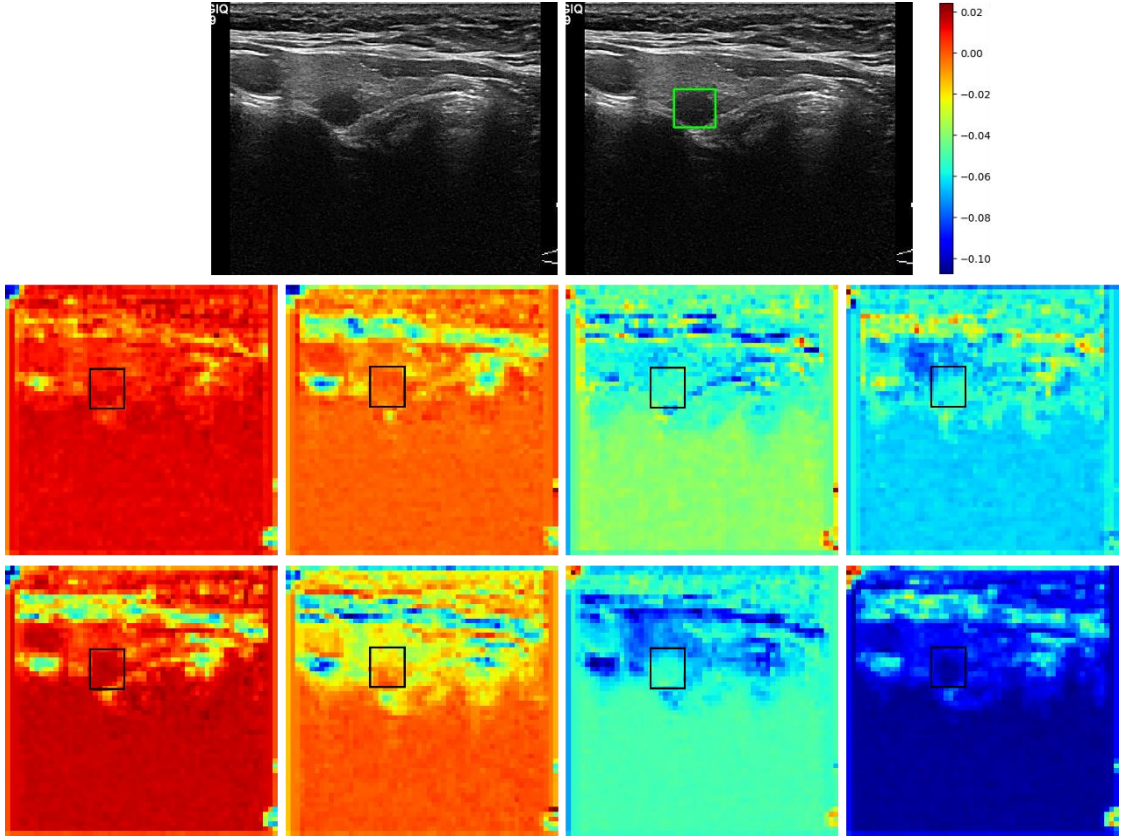

Figure. 2. Comparative visualization of feature attention maps by the MSFCA module. Top row: Input ultrasound thyroid image and its corresponding ground truth. Middle row: Feature maps from the baseline model, demonstrating scattered attention. Bottom Row: Feature maps from the MSFCA-equipped model.

MSFCA's core principle is to analyze channel attention via the frequency domain, using 2D Discrete Cosine Transform (DCT) to decompose features into various frequency bands, thereby incorporating diverse frequency components for effective channel information compression. Differing from methods reliant on only Global Average Pooling or limited frequencies, MSFCA captures richer frequency characteristics. Its key "multi-spectral" aspect enables different channel groups to focus on distinct, pre-selected frequency components. This allows the model to process a spectrum of information concurrently—from low-frequency global structures to high-frequency textural details and edges—which is vital for accurate nodule characterization. The MSFCA architecture is depicted in Figure 3.

Specifically, the input vector $X$ is firstly divided into $n$ parts along the channel dimension $[X^0, X^1, ..., X^{n-1}]$, where $X^i \in R^{C_0 \times H \times W}$, $C_0 = \frac{C}{n}$. For each part $X^i$, its 2D DCT frequency component $B_{h,w}^{u_i,v_i}$ is calculated and used as a preprocessing result for channel attention:

$$Freq^i = 2DDCT^{u_i,v_i}(X^i) = \sum_{h=0}^{H-1}\sum_{w=0}^{W-1} X_{:,h,w}^i B_{h,w}^{u_i,v_i} \quad (1)$$

$$s.t.\ i \in \{0, 1, ..., n-1\}$$

where, $u_i, v_i$ refers to 2D exponent of the frequency component to $X^i$, and $Freq^i \in R^{C_0}$ is compressed $C_0$-dimension vector after the compression. These group-specific frequency vectors are concatenated to form a C-dimensional multi-spectral vector $Freq$:

$$Freq = compress(X) = contact(Freq^0, Freq^1, ..., Freq^{n-1}) \quad (2)$$

This vector, encoding diverse frequency information, is then passed through a fully connected (*fc*) layer and a sigmoid activation to produce the channel attention weights:

$$ms_{att} = sigmoid(fc(Freq)) \tag{3}$$

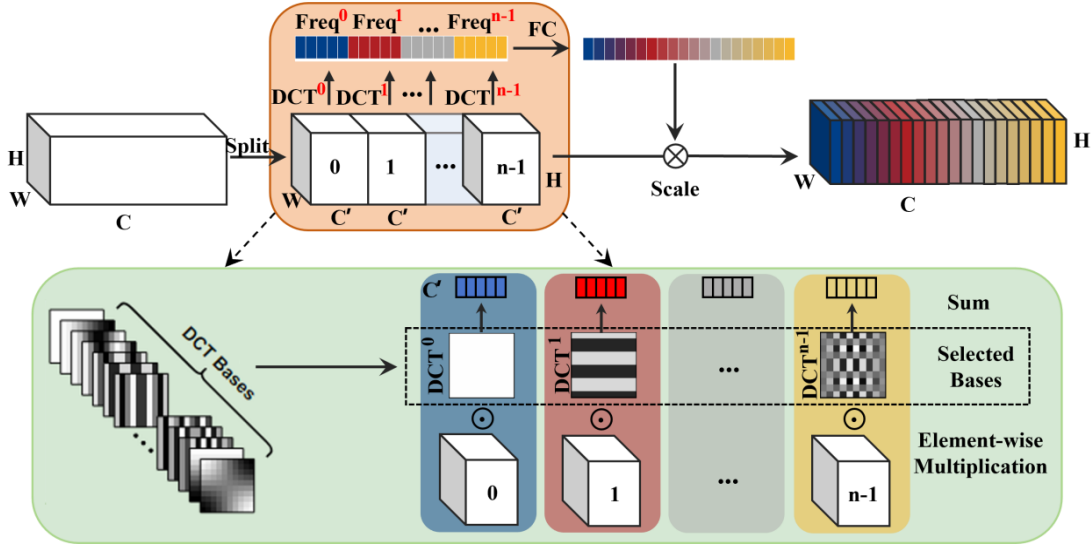

Figure.3. Multi-Spectral Frequency Channel Attention module.

In Nodule-DETR (Figure 1), MSFCA modules are integrated after each ResNet50 S1-S4 Residual block. This allows the network to adaptively recalibrate channel features using frequency-domain insights at multiple scales, improving its ability to detect challenging nodules in ultrasound images.

### 3.3. Efficient Hierarchical Feature Fusion (HFF) with Lightweight Convolutions

To enhance Nodule-DETR's performance, this paper introduces the Hierarchical Feature Fusion strategy, which refines feature fusion through an innovative architectural design (see Figures 4). The HFF approach centers on a dedicated module positioned between the ResNet50 backbone output and the deformable encoder. This module aggregates multi-scale feature maps from stages 2–4 of the backbone, processing them through channel projection to generate feature layers $P_1$, $P_2$, and $P_3$. Simultaneously, stage 4's output undergoes a 3×3 convolution and GroupNorm to generate $P_4$. These four feature layers are then integrated via a down-top pathway, where $F_1$ is derived directly from $P_1$, and each subsequent $F_i$ incorporates details from the preceding finer-grained level. This design enriches high-level semantic features with fine-grained spatial details, fostering a robust multi-scale contextual representation.

A notable advancement in the HFF framework is the Spatial-Channel Decoupling module. This module streamlines feature extraction by decoupling spatial and channel operations. It employs pointwise (1×1) convolutions to manage channel dimensions, followed by depthwise convolutions for spatial downsampling. This hierarchical fusion process can be formally expressed as:

$$F_i = \begin{cases} P_1 & i = 1 \\ P_i + SCDown(F_{i-1}) & i \in \{2, 3, 4\} \end{cases} \tag{4}$$

By incorporating the HFF strategy, Nodule-DETR more effectively preserves both high-resolution details (vital for small nodules) and rich semantic context. This enhances the model's generalization across different nodule scales, and leads to a notable performance improvement in thyroid nodule detection.

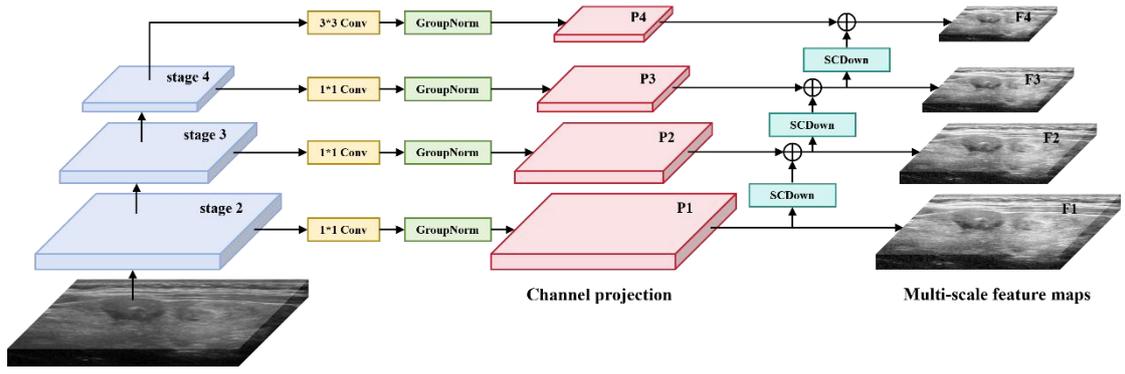

Figure.4. Hierarchical Feature Fusion module.

### 3.4. Adapting to Irregular Nodules with Multi-Scale Deformable Attention (MSDA)

Accurate detection of thyroid nodules in ultrasound images is often challenged by their small size, irregular shapes, and ambiguous boundaries, which can blend with surrounding tissues. Traditional attention mechanisms in DETR-based architectures, which process every point in a feature map, are computationally intensive and often struggle to capture the fine-grained details of such nodules. To address these challenges, we introduce the Multi-Scale Deformable Attention module, inspired by the Deformable-DETR framework [45]. This module enhances Nodule-DETR's ability to dynamically focus on critical regions, improving detection robustness for small and irregularly shaped nodules while maintaining computational efficiency. The MSDA architecture is illustrated in Figure 5.

The MSDA module builds on the concept of deformable attention, which prioritizes a small, learnable set of sampling points around a reference point, rather than processing the entire feature map. This approach allows the model to concentrate computational resources on the most informative regions, adapting flexibly to the diverse morphologies of thyroid nodules. By assigning a limited number of keys to each query, MSDA reduces computational overhead and accelerates convergence, making it particularly effective for detecting small or irregularly shaped objects. In Nodule-DETR, MSDA is seamlessly integrated into both the Transformer encoder and decoder, replacing the standard self-attention and cross-attention mechanisms, as shown in Figure 1.

The operation of a single deformable attention head is defined as:

$$\text{DeformAttn}(z_q, p_q, x) = \sum_{h=1}^{H} W_h \left[ \sum_{k=1}^{K} A_{hqk} \cdot W_m x(p_q + \Delta p_{hqk}) \right] \quad (5)$$

where $z_q$ is the query feature, $p_q$ is the 2D reference point, and x is the input feature map. The module learns sampling offsets ($\Delta p_{hqk}$) and attention weights ($A_{hqk}$) from the query feature itself. Our MSDA module extends this by operating across multiple feature levels, making it robust to variations in nodule size. The formulation becomes:

$$\text{MSDeformAttn}\left(z_q, \widehat{p_q}, \{x^l\}_{l=1}^{L}\right) = \sum_{h=1}^{H} W_h \left[ \sum_{l=1}^{L} \sum_{k=1}^{K} A_{mlqk} \cdot W_m x^l \left( \phi_l(\widehat{p_q}) + \Delta p_{mlqk} \right) \right] \quad (6)$$

Here, the attention mechanism aggregates features from L different feature map levels, with $\phi_l(\widehat{p_q})$ mapping the reference point to the l-th feature map. Based on our experiments, we set the number of feature levels L=4, the number of attention heads H=8, and the number of sampling keys K=4. Through comprehensive ablation tests, we determined that a configuration with a six-layer encoder and a six-layer decoder yields optimal performance for the Nodule-DETR model.

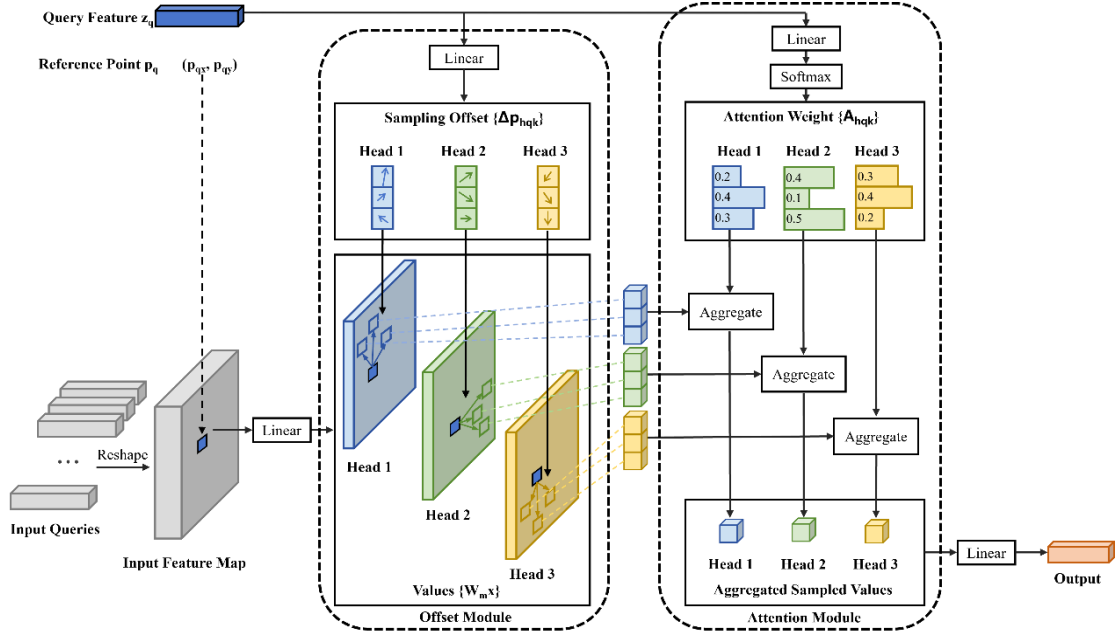

Figure.5. Deformable Attention module.

### 3.5. Loss function

During training, the model uses Hungarian loss for set prediction, ensuring that each prediction corresponds to a ground truth object with minimal cost. This assignment mechanism is defined as $\mathcal{A}_{o2o}$. The model utilizes Hungarian matching to optimize the pairing prediction between predictions and ground truths by minimizing the aggregated cost function composed of multiple loss components. The overall loss function integrates two main components: classification loss and regression loss. For classification loss, we adopt focal loss to address class imbalance, while regression loss includes L1 loss for bounding box regression and precise localization and GIoU (Generalized Intersection over Union) loss for refining spatial accuracy. The loss functions are defined below:

$$L_{loss} = L_{set}(\mathcal{A}_{o2o}(P, G)) \tag{7}$$

$$L_{set} = \lambda_{focal}L_{focal} + \lambda_{L1}L_{L1} + \lambda_{GIoU}L_{GIoU} \tag{8}$$

$$L_{focal} = -(1 - p_t)^\gamma \log(p_t) \tag{9}$$

$$L_{L1} = \frac{\sum_{i=1}^{n}|f(x_i) - y_i|}{n} \tag{10}$$

$$L_{GIoU} = 1 - \frac{|f(x_i) \cap y_i|}{|f(x_i) \cup y_i|} + \frac{|A/(f(x_i) \cup y_i)|}{|A|} \tag{11}$$

Here, $P$ and $G$ represent the sets of predictions and ground truths respectively, $\lambda_{focal}$, $\lambda_{L1}$ and $\lambda_{GIoU}$ are weight coefficients of focal loss, L1 loss and GIoU loss, which are set to 2, 5 and 2, respectively. $p_t = \begin{cases} p & y = 1 \\ 1 - p & y = -1 \end{cases}$, the value of $y$ is either 1 or -1, representing foreground and background respectively; $p$ ranges from 0 to 1, indicating the probability that the model predicts the $i$-th object as foreground, the value of $\gamma$ is 2, $f(x_i)$ and $y_i$ represent the prediction and ground truth bounding boxes of the $i$-th sample respectively, and $n$ is the total number of samples. A denotes the closure of regions $f(x_i)$ and $y_i$, which refers to the smallest axis-aligned rectangle enclosing both regions.

# 4. Experiments

## 4.1 Dataset

This study utilized a non-public dataset of 7301 thyroid nodule ultrasound images, collected from 555 videos of 201 patients at the Cancer Hospital of the Chinese Academy of Medical Sciences. The research protocol was formally approved by the Ethics Committee of the National Cancer Center (Approval No.: NCC-014607). The dataset comprises 6646 malignant and 1159 benign nodules with ROIs annotated by experienced sonographers, reflecting a common class imbalance in real-world medical data. It was divided into training, validation, and test sets with a 4089:1753:1459 ratio. Images were acquired with a GE Healthcare Logiq E11 ultrasound device and an ML6-15 (4.0-15.0 MHz) linear array probe. Figure 6 provides representative examples of benign and malignant nodules.

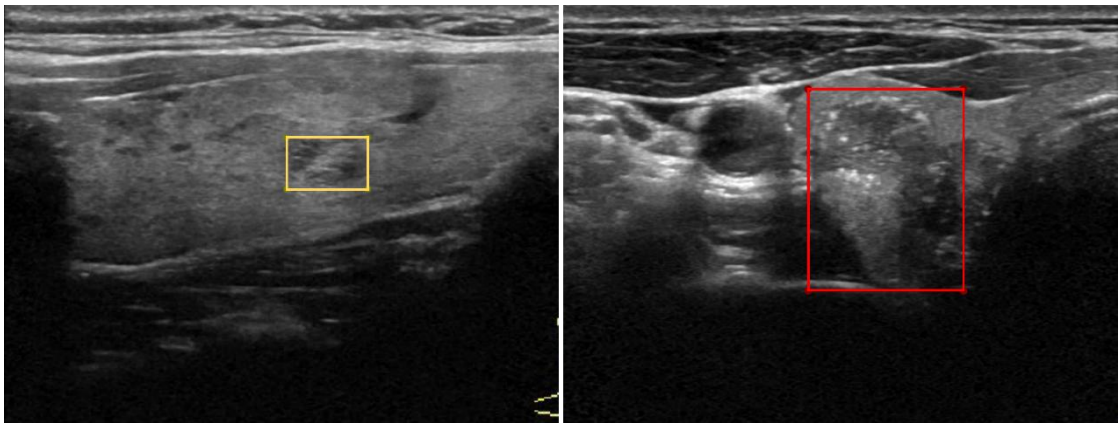

Figure. 6. Representative examples of benign(left) and malignant(right) thyroid nodules.

## 4.2. Implementation details and evaluation metrics

### 4.2.1 Implementation details

All experiments were performed on a computing environment utilizing an Ubuntu 18.04.6 LTS operating system, an Intel(R) Xeon(R) Gold 5118 CPU, 128 GB of RAM, and an Nvidia RTX 3090 GPU with 24 GB of memory. The proposed Nodule-DETR framework was implemented using Python 3.8. For the core architecture, ResNet50 was employed as the backbone network for feature extraction. The Transformer architecture consisted of six encoder layers and six decoder layers, a configuration chosen based on ablation studies that demonstrated optimal performance with this depth. It is worth noting that the NVIDIA GeForce RTX 3090 GPU used in the experiment restricted the model's depth. To avoid out-of-memory issues, both the encoder and decoder layers were limited to a maximum of 6. To improve the model's robustness and address the challenges of ambiguous nodule boundaries in ultrasound images, random noise was strategically added to both their bounding boxes and class labels during training. The hyperparameters governing this noise addition were set to 0.4 and 0.2.

Positional encoding is critical in Transformer-based models for retaining spatial information. The temperature hyperparameter for positional encoding was set to 20, a value determined through ablation experiments (see Section 4.5.3) to achieve the best balance in capturing both short-range and long-range dependencies within the image. The positional encoding was calculated using sine and cosine functions as follows:

$$\text{PE}_{(pos,2i)} = \sin(\frac{pos}{T \cdot 10000^{\frac{2i}{d_{model}}}}) \quad (12)$$

$$\text{PE}_{(pos,2i+1)} = \cos(\frac{pos}{T \cdot 10000^{\frac{2i+1}{d_{model}}}}) \quad (13)$$

where pos represents the position vector, $i$ denotes the dimension index, and $d_{model}$ refers to the feature dimension of position encoding, which is set to 64 in our experiment.

**4.2.2 Evaluation metrics**

To comprehensively evaluate the effectiveness of the nodule detection model, we use six general object detection evaluation metrics: mAP@0.5:0.95, mAP@0.5, mAP@0.75, $AP_s$, $AP_m$ and $AP_l$. Here's a breakdown of these metrics:

AP (Average Precision): This metric represents the area under the precision-recall curve, providing a measure of the model's accuracy across different confidence thresholds.

mAP (mean Average Precision): This is the average of the AP values across all object categories, giving an overall performance measure.

mAP@0.5/0.75: These variants of mAP are calculated with a specific Intersection over Union (IoU) threshold. An IoU threshold of 0.5 or 0.75 indicates that a detection is considered correct if the predicted bounding box overlaps the ground truth bounding box by at least 50% or 75%, respectively.

mAP@0.5:0.95: This metric provides a more comprehensive evaluation by averaging the AP values across a range of IoU thresholds, from 0.5 to 0.95, in steps of 0.05.

In addition to the general metrics, we also used $AP_s$, $AP_m$ and $AP_l$ to evaluate the model's detection performance with respect to different nodule sizes. Specifically, $AP_s$ evaluates the model's detection performance for small objects (area<32×32 pixels), $AP_m$ evaluates the model's detection performance for medium-sized objects (32×32<area<64×64 pixels), and $AP_l$ evaluates the model's detection performance for large objects (area > 64×64 pixels). The specific formula for mAP can be expressed as:

$$mAP = \frac{\sum_{i=1}^{N} AP(i)}{N} \quad (14)$$

where N is the total number of categories.

**4.3. Comparison with different detection algorithms**

To validate our model, we conducted a comprehensive performance comparison against a range of leading object detection algorithms. The evaluation was split into two parts: a benchmark against general state-of-the-art (SOTA) detectors and a direct comparison with methods specifically designed for thyroid nodule detection. As shown in Table 1, our Nodule-DETR model achieves the best overall performance on the thyroid nodule ultrasound dataset.

We benchmarked Nodule-DETR against several prominent models, including Faster-RCNN [33], RetinaNet [18], YOLOv7 [49], YOLOv11 [50], DETR [17], Deformable-DETR [45], DAB-DETR [44], DN-DETR [46]. While each model has its strengths, they also possess limitations for this specific nodule detection task. CNN-based models like Faster-RCNN and RetinaNet are constrained by their reliance on predefined anchor boxes, and the YOLO series, though fast, lacks robust long-range dependency modeling. Earlier Transformer-based models like DETR suffer from slow convergence. Among the more advanced models, DN-DETR stood out by introducing a novel label denoising method, achieving superior performance over its predecessors. Compared with the DN-

DETR, the proposed Nodule-DETR improves mAP@0.5:0.95, mAP@0.5, mAP@0.75, APs, APm, and APl by 0.149, 0.126, 0.231, 0.091, 0.143, and 0.215, respectively. The overall performance of the model is enhanced in thyroid nodule detection, especially in the detection of small-sized early nodules.

To further assess our model's capabilities, we also compared it against algorithms developed specifically for thyroid nodule detection by Liu et al. [34] and Zhou et al. [15]. To ensure a fair comparison, we reimplemented, and retrained these models on our dataset following the methodologies described in their respective papers. The findings confirm that our Nodule-DETR method achieves superior detection accuracy compared to both of these established, specialized models.

Table 1 Comparison of different object detection models.

| Method | Source | mAP↑ | mAP@0.5↑ | mAP@0.75↑ | $AP_s$↑ | $AP_m$↑ | $AP_l$↑ |
| --- | --- | --- | --- | --- | --- | --- | --- |
| Faster-RCNN | TPAMI2016 | 0.424 | 0.848 | 0.339 | 0.105 | 0.419 | 0.596 |
| RetinaNet | ICCV2017 | 0.407 | 0.802 | 0.353 | 0.267 | 0.418 | 0.464 |
| YOLOv7 | CVPR2023 | 0.546 | 0.883 | 0.598 | 0.308 | 0.559 | 0.638 |
| YOLOv11 | CVPR2024 | 0.584 | 0.908 | 0.665 | 0.414 | 0.581 | 0.684 |
| DETR | ECCV2020 | 0.423 | 0.823 | 0.386 | 0.216 | 0.421 | 0.566 |
| Deformable-DETR | ICLR2021 | 0.475 | 0.811 | 0.499 | 0.355 | 0.47 | 0.57 |
| DAB-DETR | ICLR2022 | 0.468 | 0.828 | 0.463 | 0.342 | 0.459 | 0.576 |
| DN-DETR | CVPR2022 | 0.462 | 0.822 | 0.481 | 0.324 | 0.479 | 0.478 |
| Liu et al. | MIA2019 | 0.558 | 0.915 | 0.613 | 0.391 | 0.568 | 0.627 |
| Zhou et al. | BSPC2024 | 0.553 | 0.904 | 0.619 | 0.345 | 0.559 | 0.628 |
| **Nodule-DETR(Ours)** | | **0.611** | **0.948** | **0.712** | **0.415** | **0.622** | **0.693** |

### 4.4. Ablation studies

To rigorously evaluate the individual contributions of the proposed architectural innovations and key hyperparameter choices within Nodule-DETR, a series of ablation studies were conducted. These experiments systematically isolate the impact of MSDA, MSFCA, HFF, the number of transformer encoder/decoder layers, and the temperature coefficient for positional encoding.

### 4.4.1. Ablation experiment of different improved methods

This section details ablation experiments on the thyroid ultrasound dataset to evaluate the individual and combined contributions of the proposed MSDA, MSFCA, and HFF. Table 2 summarizes the results and key findings are as follows.

Initially, introducing MSDA to the baseline model, which replaced original self-attention and cross-attention mechanisms in the encoder and decoder layers, substantially improved detection performance, particularly for APl and mAP@0.75. These results, detailed in Table 2, indicate that MSDA can flexibly adjust sampling regions to adapt to irregular nodules, significantly improving detection performance.

Next, MSFCA was added to the model already containing MSDA. This addition aimed to capture crucial semantic information in the frequency domain of ultrasound images. As shown in Table 2, this combination further enhanced several metrics, including mAP@0.5 and APs, highlighting the essential role of MSFCA in boosting the model's detection capabilities.

Finally, HFF based on SCDown was integrated. This involved replacing standard 3×3 convolutions in stage 3 of the backbone with lightweight SCDown and applying SCDown for feature fusion from the backbone's output. With the inclusion of HFF on top of MSDA and MSFCA, the model achieved further increases across most metrics, notably in mAP@0.75, APm, and APl, as presented in Table 2. Therefore, the observed improvements in these key areas indicate a positive overall impact on nodule detection.

Table 2 Ablation study of the proposed modules.

| MSDA | MSFCA | HFF | mAP↑ | mAP@0.5↑ | mAP@0.75↑ | $AP_s$↑ | $AP_m$↑ | $AP_l$↑ |
|---|---|---|---|---|---|---|---|---|
| | | | 0.462 | 0.822 | 0.481 | 0.324 | 0.479 | 0.478 |
| √ | | | 0.579 | 0.926 | 0.640 | 0.397 | 0.581 | 0.687 |
| √ | √ | | 0.581 | 0.946 | 0.645 | **0.416** | 0.588 | 0.662 |
| √ | √ | √ | **0.611** | **0.948** | **0.712** | 0.415 | **0.622** | **0.693** |

### 4.4.2. The comparison between the number of encoder and decoder layers

The depth of the Transformer architecture plays a critical role in model performance. The encoder is responsible for learning global while the decoder models the interactions among the various feature representations to generate predictions. To analyze the impact of model depth, we conducted ablation studies by varying the number of encoder and decoder layers.

The results, presented in Table 3, show that model performance generally improves as the number of encoder layers is increased from zero to six. This trend underscores the importance of deep global feature learning for accurately identifying thyroid nodules and their spatial configurations. Likewise, Table 4 demonstrates that detection performance consistently improves as the number of decoder layers increases, with the six-layer configuration achieving the best results.

Our ablation tests identified a six-layer encoder and six-layer decoder as the optimal configuration. It is crucial to note that this maximum depth was determined by the hardware constraints of our experimental setup. The model was trained on an NVIDIA GeForce RTX 3090 GPU, and to avoid out-of-memory issues, both the encoder and decoder were limited to a maximum of six layers. Thus, the chosen architecture represents the best-performing configuration achievable within the memory limits of the device.

Table 3 Ablation study with different layers of encoders.

| Encoder layers | mAP↑ | mAP@0.5↑ | mAP@0.75↑ | $AP_s$↑ | $AP_m$↑ | $AP_l$↑ |
|---|---|---|---|---|---|---|
| 0 | 0.575 | 0.942 | 0.614 | 0.389 | 0.581 | 0.658 |
| 1 | 0.593 | 0.940 | 0.659 | 0.402 | 0.597 | **0.693** |
| 3 | 0.603 | **0.950** | 0.679 | **0.418** | 0.612 | 0.688 |
| 6 | **0.611** | 0.948 | **0.712** | 0.415 | **0.622** | **0.693** |

Table 4 Ablation study with different layers of decoders.

| Decoder layers | mAP↑ | mAP@0.5↑ | mAP@0.75↑ | $AP_s$↑ | $AP_m$↑ | $AP_l$↑ |
|---|---|---|---|---|---|---|
| 1 | 0.435 | 0.766 | 0.441 | 0.205 | 0.445 | 0.541 |
| 3 | 0.598 | 0.945 | 0.663 | 0.382 | 0.604 | 0.691 |
| 6 | **0.611** | **0.948** | **0.712** | **0.415** | **0.622** | **0.693** |

### 4.4.3. Hyperparameter tuning of temperature for positional encoding

This section examines the impact of the temperature coefficient in Transformer layer positional encoding on DETR's sensitivity to positional information. The temperature coefficient directly influences the frequency of sine and cosine functions in positional encoding, thereby influencing the model's ability to capture position-related dependencies. A higher temperature coefficient increases sensitivity to position, while a lower one emphasizes global context. Our experiments (Table 5) show that a temperature coefficient of 20 provides the best balance for capturing both short-range and long-range dependencies, leading to optimal performance across various AP metrics. We hypothesize that lower frequencies (e.g., temperature=30) might over-smooth positional encoding, hindering short-range dependency learning, while higher frequencies (e.g., temperature=1 or 10) could cause abrupt changes, making long-range dependency learning difficult.

Table 5 Ablation study of temperature for positional encoding.

| temperature | mAP↑ | mAP@0.5↑ | mAP@0.75↑ | $AP_s$↑ | $AP_m$↑ | $AP_l$↑ |
|---|---|---|---|---|---|---|
| 1 | 0.579 | 0.930 | 0.641 | **0.415** | 0.580 | 0.675 |
| 10 | 0.580 | 0.941 | 0.640 | 0.378 | 0.588 | 0.671 |
| 20 | **0.611** | **0.948** | **0.712** | **0.415** | **0.622** | **0.693** |
| 30 | 0.581 | 0.942 | 0645 | 0.380 | 0.590 | 0.668 |

### 4.5. Qualitative Analysis and Visualization

To provide a qualitative assessment of our model's performance, we visualized the detection results from Nodule-DETR and compared them against those from other leading algorithms on a set of randomly selected ultrasound images. As shown in Figure 7, these examples include challenging cases with both benign and malignant nodules.

The visualization highlights the limitations of other existing models when applied to this task. The CNN-based detectors exhibited several issues. While Faster-RCNN [42] achieved high confidence on some nodules, it also produced a clear misclassification (image 4). RetinaNet's [43] predictions generally had lower confidence scores than our method. The YOLO models also struggled, with YOLOv7 [44] showing over-detection on one image and YOLOv11 [45] producing duplicate detections on another. The DETR-based [17] models showed inconsistent performance. The original DETR performed poorly on benign nodules, with low confidence and over-detection (column 4 and 5). Deformable-DETR [31] missed a nodule entirely (column 4) and over-detected on another (column 5), while DAB-DETR [41] also had a missed detection (column 4). The baseline model, DN-DETR [32], performed well without errors like over-detection or misclassification, but its confidence scores were still lower than those of our proposed method. The specialized method from Liu et al. [34] exhibited over-detection on two images , and while the method from Zhou et al. [15] was accurate, its overall confidence was lower than Nodule-DETR's.

In contrast, our Nodule-DETR model accurately predicts both the location and the class of thyroid nodules with high confidence scores. The visual evidence in Figure 7 shows that our method avoids the missed detections, misclassifications, and over-detections observed in the other models. Overall, the qualitative results strongly support the quantitative findings, demonstrating that Nodule-DETR is a more accurate and reliable detection model for both common and small, early-stage thyroid nodules, making it a valuable tool to support clinical diagnosis.

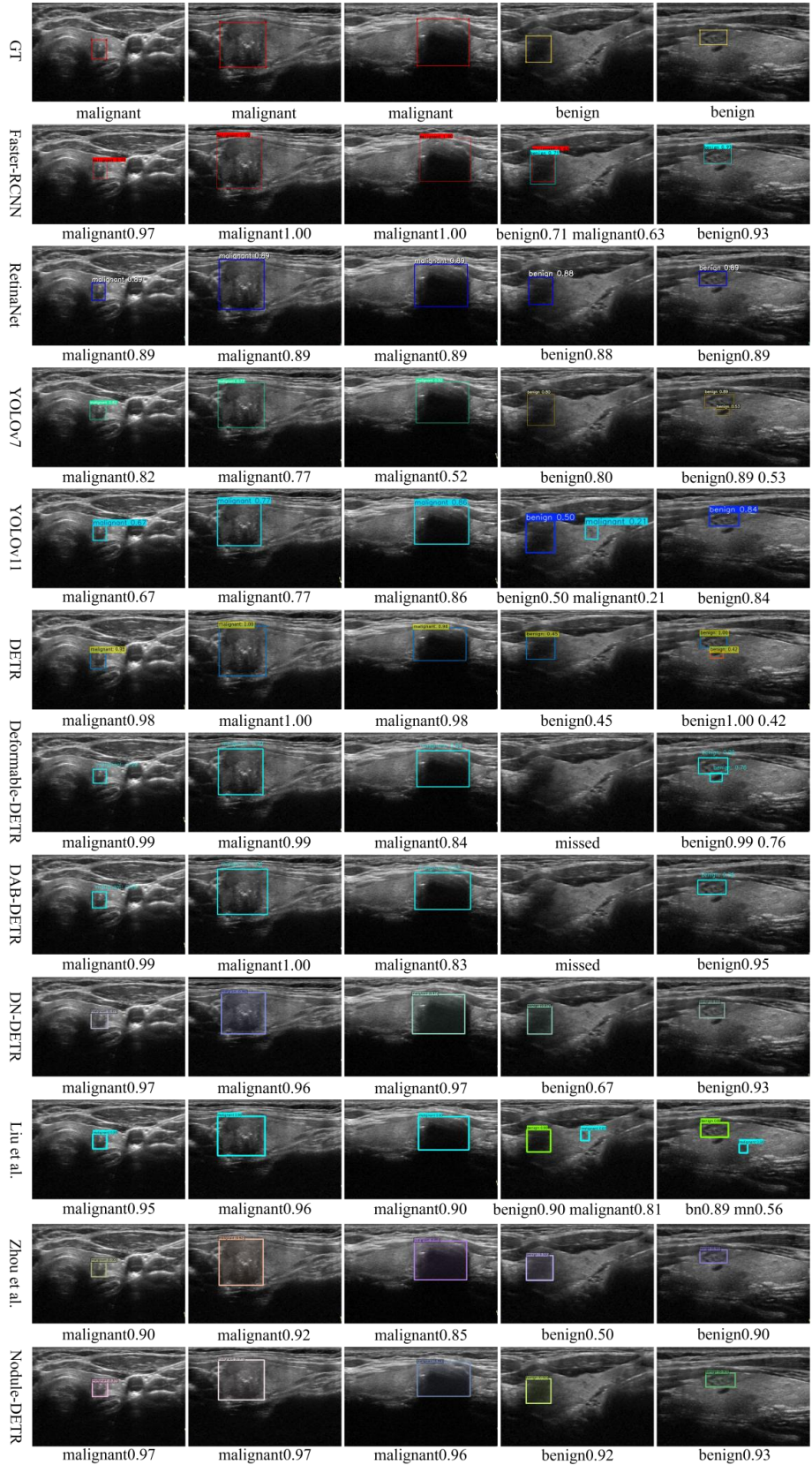

Figure. 7. Qualitative comparison of thyroid nodule detection performance. The top row (GT) shows ground truth annotation and the classification. The number following each detection is the confidence score of the classifications of benign and malignant nodules.

# 5. Discussion and Conclusion

## 5.1. Discussion

Early and accurate diagnosis of thyroid nodules is of great significance for the prevention and treatment of thyroid cancer. While ultrasound examination is the primary method for diagnosing thyroid nodules, there are still significant challenges in distinguishing between benign and malignant nodules. Current diagnostic workflows depend on radiologists' expertise and subjective interpretation, introducing inter-observer variability that not only increases physician workload but also compromises diagnostic accuracy. Therefore, it is necessary to develop automatic diagnostic assistance systems.

To better support the clinical diagnostic decisions of radiologists, we have proposed a deep learning-based detection method, Nodule-DETR, for the automated detection and classification of thyroid nodules. The superior performance of the Nodule-DETR model is attributable to three key contributions. First, the integration of a Multi-Spectral Frequency Channel Attention module into the feature extraction network enhances the model's ability to detect nodules with low contrast and blurred boundaries. Second, a Hierarchical Feature Fusion module is designed to improve multi-scale feature fusion, allowing the model to capture detailed feature information often overlooked while reducing interference from surrounding tissue and redundant information. Third, Multi-Scale Deformable Attention is incorporated into the encoder and decoder, granting the network greater flexibility in capturing the features of small and irregularly shaped nodules. As validated by the ablation studies (Table 2), each proposed component contributes to a progressive improvement in performance metrics. Furthermore, comparative experiments detailed in Table 1 confirm that the proposed Nodule-DETR method demonstrates superior performance against other state-of-the-art detection models.

Despite the promising results, we acknowledge some limitations in this study. A primary limitation is that our model, like many current computer-aided diagnosis systems, operates on static 2D ultrasound images. This approach does not capture the dynamic and temporal information available in ultrasound video sequences. Future work will focus on extending our current model from static image analysis to dynamic video analysis to achieve a more comprehensive and robust diagnostic assessment, more closely mirroring the clinical practice of experienced sonographers. Another limitation is the current reliance on extensive, manually annotated datasets, a process which is both time-consuming and labor-intensive for expert radiologists. To mitigate this, we plan to investigate the use of weakly-supervised or unsupervised learning techniques. This approach, combined with data collection from multiple centers, could reduce the dependency on manual annotations and improve the model's generalizability.

## 5.2. Conclusion

In this study, we propose a DETR-based detection model, Nodule-DETR, for the automatic detection and classification of thyroid nodules in ultrasound images. Firstly, a Multi-Spectral Frequency Channel Attention is introduced into the CNN backbone to enhance the model's ability to detect nodules with low contrast and blurred boundaries. Secondly, a Hierarchical Feature Fusion module is designed between the backbone and the deformable encoder to efficiently fuse multi-scale features, which improves the multi-scale ability of the model. Furthermore, Multi-Scale Deformable Attention is added to the encoder and decoder to better capture the shape information of small and irregularly shaped nodules, further improving the detection ability of the model for irregularly shaped nodules. Experimental results show that our method achieves superior detection results and higher accuracy for thyroid nodules in ultrasound images compared to other existing methods. This indicates that Nodule-DETR can effectively reduce the workload of radiologists and holds important clinical application potential.

## Acknowledgement

This work was supported by Beijing Natural Science Foundation under Grant L242112, Grant L222034, Grant L222104, and Grant L232037; in part by the National Natural Science Foundation of China under Grant 12375359.